# Enhancing Learnability of classification algorithms using simple data preprocessing in fMRI scans of Alzheimer's disease

Rishu Garg[1], Rekh Ram Janghel[2] and Yogesh Rathore[2]

[1, 2, 3] National Institute of Technology, Raipur, India

[1] rishu.garg.rg06@gmail.com, [2] rrjanghel.it@nitrr.ac.in, [3] yogshrathore23@gmail.com

**Abstract.** Alzheimer's Disease (AD) is the most common type of dementia. In all leading countries, it is one of the primary reasons of death in senior citizens. Currently, it is diagnosed by calculating the MSME score and by the manual study of MRI Scan. Also, different machine learning methods are utilized for automatic diagnosis but existing has some limitations in terms of accuracy. In this paper, we have proposed some novel preprocessing techniques that have significantly increased the accuracy and at the same time decreased the training time of various classification algorithms. First, we have converted the ADNI dataset which was in 4D format into 2D form. We have also mitigated the computation costs by reducing the parameters of the input dataset while preserving important and relevant data. We have achieved this by using different preprocessing steps like grayscale image conversion, Histogram equalization and selective clipping of dataset. We observed a highest accuracy of 97.52% and a sensitivity of 97.6% in our testing dataset.

## 1 INTRODUCTION

Alzheimer's disease is a neurodegenerative disease that leads to gradual memory loss in human beings. In most cases, Alzheimer's disease leads to dementia [1]. Although there is no definite and effective cure for Alzheimer's disease, early detection and treatment can reduce the severity of symptoms which can alleviate the sufferings of the patient. Medications and treatments are most effective when the disease is detected in its early stages. For instance, early treatment can slow down memory loss for some period. [2]. Alzheimer's report two common abnormalities in the brain of this patient, "1. Dense layers of protein deposited outside and between the nerve cells. 2. Areas of damaged nerve fibers, inside the nerve cells, which instead of being directly had become tangled". Moreover, these plaques and tangles have been used to help diagnose AD [3]. To overcome these problems, Diagnosing Alzheimer's needs careful medical assessment, as well as patient records, mental state examination (MMSE), and many neurobiological and physical examinations[4]. Besides, s-MRI (structural magnetic resonance imaging) and rs-fMRI (resting-state functional magnetic resonance imaging) is the most common method of analyzing the regular changes, different activities in the brain [5].

## 2     LITERATURE REVIEW

Jun Jie Ng et. al. [5], proposed a method where machine learning algorithms are used to build up knowledge of the patient's behavior over time. It was used mainly to locate the position of patients around the house via the help of Estimote Bluetooth beacons and could pinpoint which room the patient was in up to an accuracy of 95%. Lauge Sørensen et. al. [6], proposed a study, they investigate hippocampal texture as an MRI-based feature for the identification of Alzheimer's disease at an early stage. Through this study accuracy achieved at 83%. Here they found that the hippocampal texture feature had a notably superior classification between stable MCIs and MCI-to-AD converters. Siqi Liu et al. used a stacked auto-encoders of deep learning and at the output layer, they used Softmax, to reduce the bottleneck problem [7]. While comparing with other previous techniques, their method can classify data of multiple classes, needs less training data and also needs very less information about input data. They achieved a considerable performance of 87.67% accuracy on the classification of all diagnosis groups. They come with a result that, classification techniques need to combine multiple features to get more accurate classification results. Tong Tong et.al. [8], present a paper, they present a categorization structure to accurately utilize the complementarity in the different input dataset. Features from many modalities are then collected using a nonlinear graph mixture process, which produces a fused graph for final classification. Using these fused graphs, they got a classification area under the curve (AUC) of the receiver-operator attribute of 98.1% between normal controls (NC) images and AD images, 82.4% between MCI images and NC images and 77.9% in the overall classification. Tijn M. Schouten, et al. [9], uses different techniques to take out features from the diffusion Magnetic Resonance Images. First, they use the pixel-wise distribution tensor calculations that have been frame worked using region-based spatial statistics. Second, they clustered the pixel-wise distribution calculations using ICA and they calculated the fusion of these features. Here, Table 1 contains the major issues covered by different researchers on the ADNI dataset, where we include the methods applied by different authors on the same dataset and what accuracy they got after these methods.

**Table 1. Literature survey**

| Title | Author and Year | Dataset | Method | Accuracy | Sensitivity | Specificity |
|---|---|---|---|---|---|---|

| Title | Author | Dataset | Method | Accuracy | Sensitivity | Specificity |
|---|---|---|---|---|---|---|
| Convolutional neural network based Alzheimer's disease classification from magnetic resonance brain images | Rachana Jain et al (January 2019) | ADNI | VGG-16, $P_fS_eCl$ | 95.73% | - | - |
| Fast Training of a Convolutional Neural Network for Brain MRI Classification | Zongjie Tu et al (April 2019) | ADNI | Modified VoxCNN | 86.3±.061 | - | - |
| K-Means clustering and neural network for object detecting and identifying abnormality of brain tumor | N. Arunkumar et al(November 2018) | BRATS training dataset | K means Clustering, ANN, SVMs | 90.9 | - | - |
| Automated Categorization of Multi-Class Brain Abnormalities Using Decomposition Techniques With MRI Images: A Comparative Study | Anjan Gudigar(Feb 2019) | Harvard Medical School Database | Variational Mode Decomposition (VMD) | 90.68 | 99.43 | - |
| Performance Improved Iteration-Free Artificial Neural Networks for Abnormal Magnetic Resonance Brain Image Classification | D. Jude Hemanth et al(December 2011) | A set of 540 MR brain tumor images | Modified Counter Propagation Neural Network (MCPN) and Modified Kohonen Neural Network (MKNN). | 95 | 90 | 96 |
| Using Deep CNN with Data Permutation Scheme for Classification of Alzheimer's Disease in Structural Magnetic Resonance Imaging (sMRI) | Bumshik Lee et al(July 2019) | ADNI Dataset | AlexNet, Transfer Learning, CNN | 98.74 | - | - |

## 3 METHODOLOGY

### 3.1 Dataset Description

In this paper, we acquired data from Alzheimer's Disease Neurological Initiative (ADNI). The ADNI image dataset contains 1917 images belonging to AD (Alzheimer's Disease) class and 1775 images belonging to NL (Normal person) class. Thus the total images are 3692.

### 3.2 Dataset preprocessing and Training

**Image conversion:**

   Input: List of 3D Images
   Output: List of 2D Images

Begin
1. Read Input image
2. Import dicom // to read nifty images
3. Import Numpy // to modify numpy array
4. FOR EACH Image in 3D Images
5. Read Image ← dicom.load(image)
6. Image_Shape ← Image.shape()
7. Store in x ← Image_Shape[0] // Height
8. Store in y ← Image_Shape[1] // Width
9. Store in z ← Image_Shape[2] // Length
10. FOR n in RANGE 0 to z
11. Save new_image ← Image.Save(x,y)
12. Save .jpg file
13. Save new_image as Image_n // where n is a number for 0 to z
14. End FOR loop
15. End FOR EACH Loop
End

**Segmentation and gray scale conversion:**

The dataset consists of various cross-section layers of brain MRI scans of the healthy and diseased person. In this paper, we have tried to modify the dataset. For the maximum efficiency of our model, we have clipped the dataset for the cross-sectionswhich are nearer to the edge of the skull.

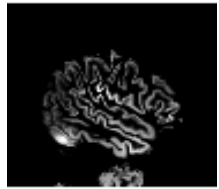

Figure-1: Cross section of Brain scan far away from the edge of the skull

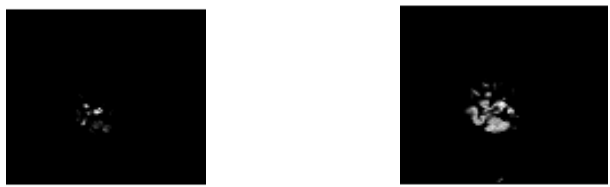

Figure-2: Cross section of Brain scans near the edge of the skull

The images located near the edge of the skull are not suitable for extracting any meaningful features that can help the network. Hence, these images were clipped from the training and testing dataset. We have further improved our dataset by modulating the pixel intensities using the Histogram Equivalence method. It helped us to intensify the overall contrast of the dataset, hence assisting in extracting the features of the images. The number of parameters in the dataset was reduced by converting the RGB images into grayscale versions. Reducing parameters won't result in any significant data loss since the images are already present in Black and White format. Since the input dataset consists of images in black and white.

The detailed methodology is described in the Figure-3 below.

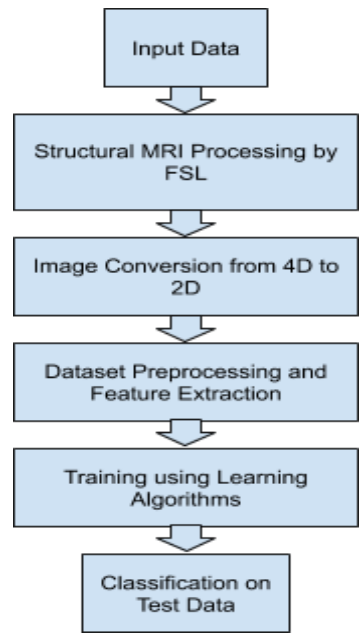

Figure-3: Figure of detailed Methodology

### 3.3    Classification

The preprocessed data is then used to train standard classification algorithms namely Random Forest Algorithm, XGBOOST classification and a simple CNN model as shown in figure-4.

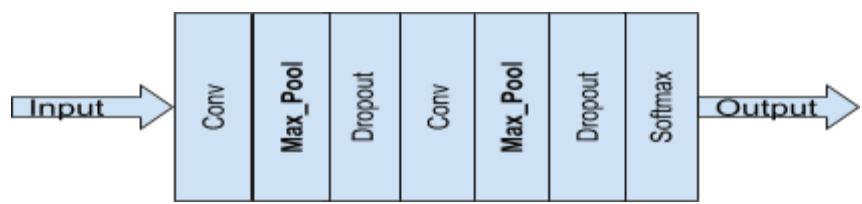

Figure-4: Architecture of proposed CNN model

## 4 RESULTS AND CONCLUSION

The models were created and trained on Kaggle kernels. The model was trained for a 40 epochs. We calculated the accuracy and sensitivity for assessing the correctness of our models. Also, the confusion matrix was calculated as shown in Figure for further assessment. The best test results produced an Accuracy of 97.52% and a sensitivity of 97.60%.The test accuracy and test loss was also plotted as shown in table 2. Table 3 presents the calculation of computation time of these algorithms.

Table 2: comparison of accuracy before and after pre processing

| Learning Algorithm | Accuracy( before preprocessing) | Accuracy(After preprocessing) |
|---|---|---|
| Random Forest | 82.81% | **86.92%** |
| XGBOOST | 90.12% | **92.05** |
| Shallow CNN | 95.13% | **97.52%** |

Table 3 comparison of computation time before and after pre processing

| Learning Algorithm | Computation Time (before preprocessing) | Computation Time (After preprocessing) | Percentage Decrease |
|---|---|---|---|
| Random Forest | 21.86 seconds | 8.66 seconds | 60.38% |
| XGBOOST | 1436.93 seconds | 378.63 seconds | 73.65% |
| Shallow CNN | 126.85 seconds | 83.72 seconds | 34% |

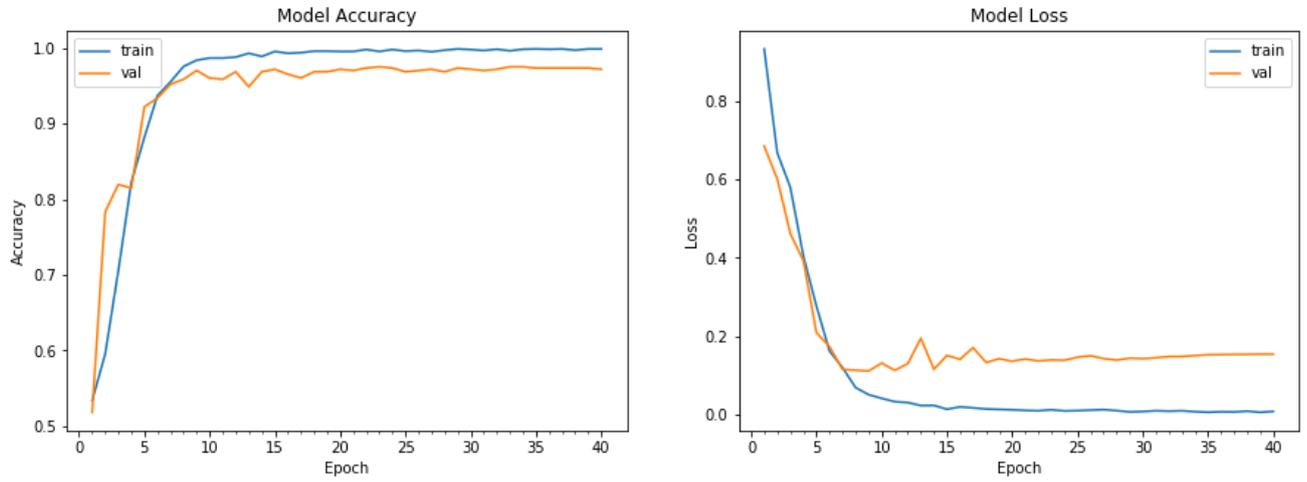

Figure-5: Accuracy and loss curve

| CNN model | XGBOOST | Random Forest |
|---|---|---|
|  |  |  |

Figure-6: Confusion Matrix of various models

From figure-5, we can observe that the accuracy of 97.52% can be achieved using simple CNN model with 2 convolution layer. So we can easily conclude that rather than applying more complex models on ADNI dataset we can go with very simple CNN model after applying strong pre-processing like histogram equalization, segmentation, type conversion etc. With the help of data preprocessing, we were able to reduce the model complexity by reducing the number of parameters and hence, with the same amount of data, we were not only able to get a better accuracy but also reduced the training time. Our method can also be used in similar applications having insufficient amount of training data.